\documentclass[letterpaper, 10 pt, conference]{ieeeconf}  %

\IEEEoverridecommandlockouts                              %

\usepackage{url}
\usepackage{graphicx}
\usepackage{subfig}
\usepackage{multirow}
\usepackage{tikz}
\usepackage{amsmath}
\usepackage{hyperref}

\def \nobreakseq {\nobreak \hskip 0pt \hbox}

\graphicspath{ {./figures/} }

\newcommand\copyrighttext{\footnotesize \textcopyright~2020 IEEE. Personal use of this material is permitted.  Permission from IEEE must be obtained for all other uses, in any current or future media, including reprinting/republishing this material for advertising or promotional purposes, creating new collective works, for resale or redistribution to servers or lists, or reuse of any copyrighted component of this work in other works.
}%

\title{\LARGE \bf Traffic Control Gesture Recognition for Autonomous Vehicles}

\usepackage{xcolor}
\author{
Julian Wiederer$^{1,2,\dagger}$, Arij Bouazizi$^{1,2,\dagger}$, Ulrich Kressel$^{1}$, Vasileios Belagiannis$^{2}$%
\thanks{$^{1}$Mercedes-Benz AG, Heßbrühlstraße 21, 70565 Stuttgart, Germany.}%
\thanks{$^{2}$Universit\"at Ulm, Albert-Einstein-Allee 41, 89081, Ulm, Germany.}
\thanks{$\dagger$ denotes equal contribution.}
\thanks{E-mail: \textit{firstname.lastname@\{daimler.com, uni-ulm.de\}}.}%
\thanks{$^{3}$ Project page: \url{https://github.com/againerju/tcg_recognition}}
}

\newcommand\copyrightnotice{%
	\begin{tikzpicture}[remember picture,overlay]
	\node[anchor=south,xshift=0pt,yshift=14pt] at (current page.south) {\fbox{\parbox{\dimexpr\textwidth-\fboxsep-\fboxrule\relax}{\copyrighttext}}};
	\end{tikzpicture}%
}

\begin{document}

\maketitle
\thispagestyle{empty}
\pagestyle{empty}

\begin{abstract}
A car driver knows how to react on the gestures of the traffic officers. Clearly, this is not the case for the autonomous vehicle, unless it has road traffic control gesture recognition functionalities. In this work, we address the limitation of the existing autonomous driving datasets to provide learning data for traffic control gesture recognition. We introduce a dataset that is based on 3D body skeleton input to perform traffic control gesture classification on every time step. Our dataset consists of 250 sequences from several actors, ranging from 16 to 90 seconds per sequence. To evaluate our dataset, we propose eight sequential processing models based on deep neural networks such as recurrent networks, attention mechanism, temporal convolutional networks and graph convolutional networks. We present an extensive evaluation and analysis of all approaches for our dataset, as well as real-world quantitative evaluation. The code and dataset is publicly available$^3$.
\end{abstract}

\section{INTRODUCTION}

\copyrightnotice

Part of autonomous driving incorporates the vehicle interaction with humans. In urban traffic situations, the interaction engages pedestrians, school traffic patrols and traffic officers among others. The latter two examples are particularly interesting for the road traffic control. While a driver has learnt to recognise the traffic hand signals, it is not the same for the autonomous vehicle. Traffic control signals, i.e. hand gestures, need to be ``taught'' to the autonomous vehicle by means of learning databases. Understanding those gestures is essential for achieving proactive and safe autonomous driving.

On one hand, recent perception databases for autonomous driving, e.g. Cityscapes~\cite{cordts2016cityscapes}, ApolloScape~\cite{huang2018apolloscape} or Eurocity Persons Dataset~\cite{Braun2019TheDetection}, contain thousands of pedestrians, road users and cyclists, however, due to the rareness of gestures they lack scenarios with human-vehicle interaction. Road traffic controllers do not exist in this kind of databases. On the other hand, gesture recognition databases~\cite{Escalera2015ChalearnResults} include body-language, as well as human-human~\cite{Joo2016PanopticCapture} and human-machine~\cite{zengeler2019hand} interactions, but they lack of road traffic control gestures, such as stop or go. It becomes, thus, a necessity to create a public database for road traffic control gesture recognition.

In this work, we introduce the TCG dataset for road traffic control gesture recognition, targeted on autonomous vehicles. We define the gestures as a set of landmarks that belong to the general body pose, represented by a three-dimensional skeleton. The aim of the dataset is to classify the traffic control gestures in every time step from the sequential skeleton-based input. With the progress in human pose estimation \cite{Belagiannis2014HolisticForests, belagiannis20153d, Belagiannis2017RecurrentEstimation} the body skeleton representation has become a standard input for gesture and activity recognition~\cite{jhuang2013towards, de2016skeleton, Nunez2017ConvolutionalRecognition}. Moreover, it allows generalization to any kind of road traffic controller since it does not depend on the individual's appearance. Capturing outdoors skeleton-based traffic control gestures is not trivial though. Motion capture on public roads is forbidden due to road obstruction. To address this limitation, we work on a closed environment where we portray road intersections with multiple vehicles and the road traffic controller involved. Our recordings include all possible traffic control scenarios for road intersections with a large amount of human motion variance. Finally, our quantitative evaluations on real-world sequences show that our studio-based recordings capture the variance of the real-world. This is the first public dataset for traffic control gesture recognition to the best of our knowledge.

\begin{figure}[t]
    \centering
    \includegraphics[width=0.95\columnwidth]{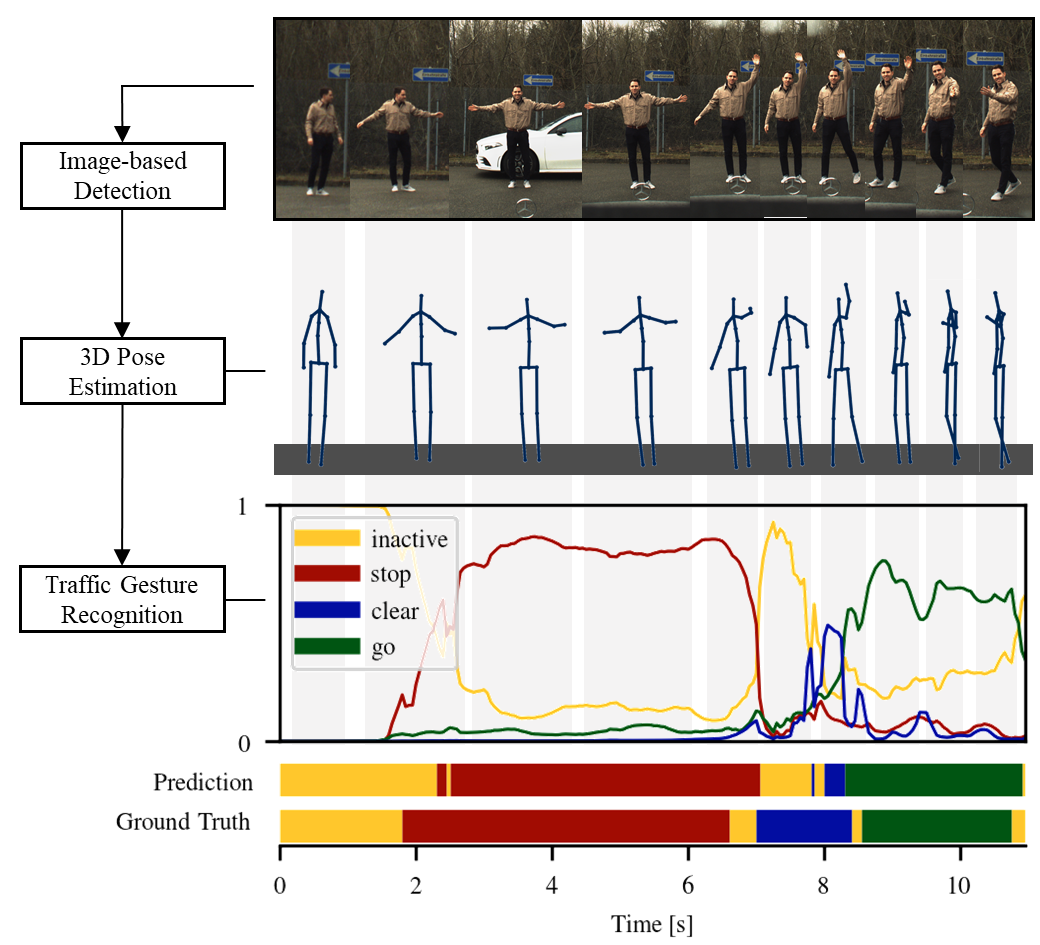}
    \caption{Real-World Results. We demonstrate how our work functions in real-world traffic control scenarios. First, we locate the traffic controller with image-based detection and 2D body pose estimation. Second, we use 3D lifting to transform the 2D to 3D pose skeleton from a sequence of estimates. Then, the temporal gesture recognition approach predicts the gesture category based on the sequence of 3D body skeletons.}
    \label{fig:real_world_demo}
\end{figure}

Alongside with the dataset, we examine a plethora of neural network approaches for gesture recognition from sequential data. In traffic control gesture recognition, we have a sequence to sequence problem where the gesture classification happens for each input of a 3D body skeleton. This mapping is modeled with recurrent neural networks (RNNs), including attention models, temporal convolutional networks (TCNs) and graph-based networks (GCNs). In total, we examine eight different neural networks architectures, demonstrating the advantages and limitations for each model. For that reason, we provide an extensive evaluation on our dataset and real-world sequences for cross-subject and cross-view settings, using multiple metric scores. On the real-world evaluation (see Fig.~\ref{fig:real_world_demo}), we demonstrate that our dataset generalizes well outdoors, although it has been captured on a closed environment.

To sum up, our work makes the following contributions: 1.~The first public traffic control gesture recognition dataset for autonomous vehicles. 2.~An extensive evaluation of eight sequence modelling approaches, including recurrent networks, attention mechanism, TCN and GCN models. 3.~An quantitative evaluation on real-world sequences to show generalization.

\section{RELATED WORK}

Gesture recognition for human-machine and human-human interaction is a long studied problem~\cite{rautaray2015vision, pisharady2015recent}. Below, we discuss the related datasets and approaches to gesture recognition, where our focus is on human-vehicle interaction.

\noindent \textbf{Human-vehicle interaction.} Autonomous vehicles need to interact with humans inside the vehicle~\cite{pickering2007research, ohn2014hand}, e.g. driver, cyclist and passengers, as well as outside the vehicle, e.g.~pedestrians and police~\cite{rasouli2019autonomous}. According to these studies, comprehensive understanding of the body language is important in order to react according to the human intentions. In particular, hand gestures are a common mean of interaction between the vehicle and human~\cite{Sachara2017Free-handReal-time,  Zengeler2019HandCameras}. Fortunately, the state-of-the-art on gesture recognition~\cite{molchanov2015hand, de2016skeleton, lindgren2018learned} allows to make easily accurate predictions. However, modeling the traffic control gestures can be challenging due to the intercultural differences~\cite{Gupta2016ConventionalizedVehicles}. For example, the traffic control hand gestures differ from country to country. In addition, road traffic control gestures are unique and they are not included in general gesture recognition datasets. In this work, we focus on the German traffic control gestures, which are also common in Europe.

\noindent \textbf{Traffic control gesture recognition.} Although traffic control gesture recognition becomes increasingly important in autonomous driving, the prior work on the problem is rather limited. Recently, Ma \textit{et al}.~\cite{Ma2018TrafficNetwork} have developed a spatiotemporal convolutional neural network (CNN) to spot Chinese traffic command gestures. Similarly, a long short-term memory (LSTM) network is employed in \cite{He2019VisualFeatures} for classifying also Chinese traffic police gestures. Both approaches rely on human body skeleton input to perform the recognition. As the human body pose is in general a strong feature for activity recognition~\cite{jhuang2013towards, de2016skeleton}, we also build our baselines with skeleton-based input. Compared to these prior approaches, we do not only study the problem by providing a number of algorithmic solutions, motivated by general gesture recognition, but we additionally release a public database for traffic control gesture recognition.

\noindent \textbf{Existing gesture recognition databases.} A reason for the limited research on traffic control gesture recognition is due to the lack of public data. While there are several hand gesture databases~\cite{pisharady2013attention} for indoor scenarios~\cite{pisharady2015recent}, general gestures~\cite{escalera2013multi} and for specific applications such as sign language recognition~\cite{pugeault2011spelling} or egocentric gesture recognition~\cite{cao2017egocentric}; the publicly available databases for traffic control hand gesture recognition are inexcitable. Consequently, our new public database on traffic control hand gesture supports the further research on the problem. Next, we introduce our dataset and then present the baseline algorithms for evaluation. 

\section{TRAFFIC CONTROL GESTURE DATASET}

We introduce TCG, a dataset for traffic control gesture recognition, that covers all possible road traffic control variations for European road intersections. By modeling road intersections, we automatically include the non-intersection situations as well. We consider road traffic control gesture recognition as a classification task from 3D body pose skeleton input data over time. As a result, our dataset consists of 3D human body skeleton sequences represented by joint sets and the respective label per skeleton. Below, we discuss the data collection, labelling and properties, as well as the experimental setup.

\begin{figure}[t]
    \centering
    \subfloat[Stop. \label{fig:samples_stop}]
    {\includegraphics[width=0.19\columnwidth]{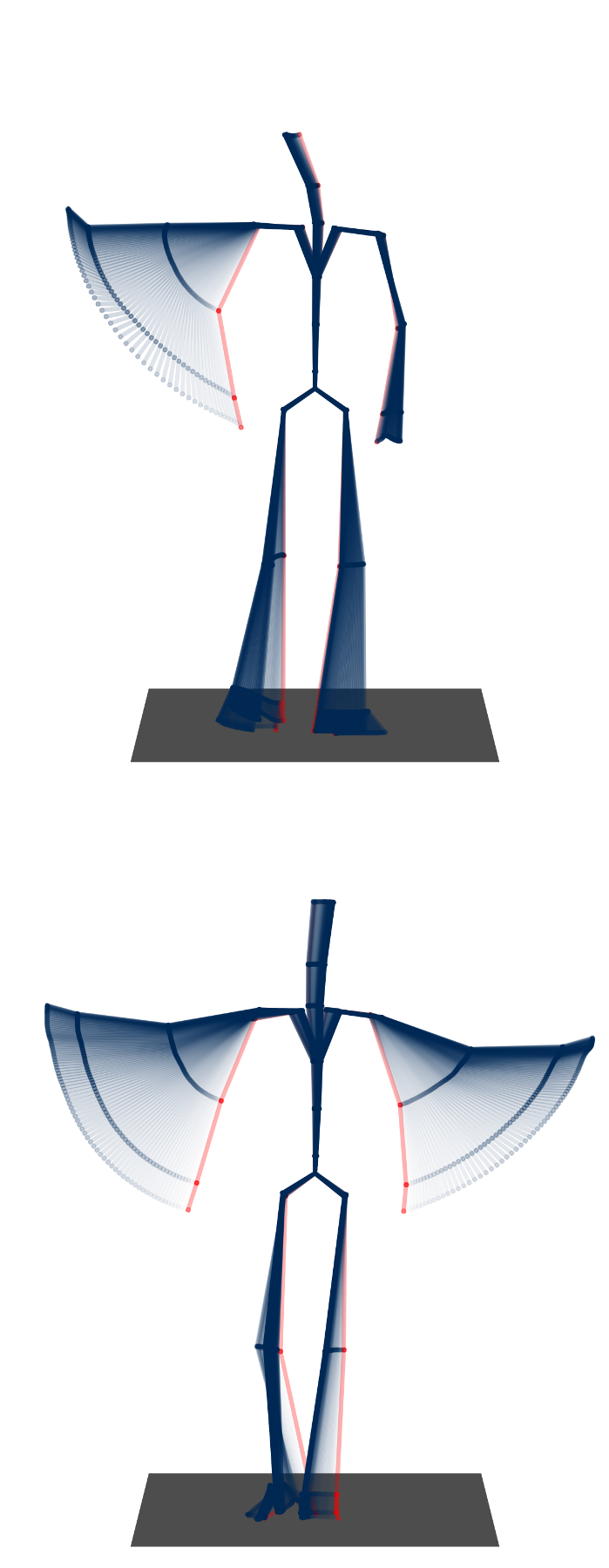}}
    \subfloat[Clear. \label{fig:samples_clear}]
    {\includegraphics[width=0.19\columnwidth]{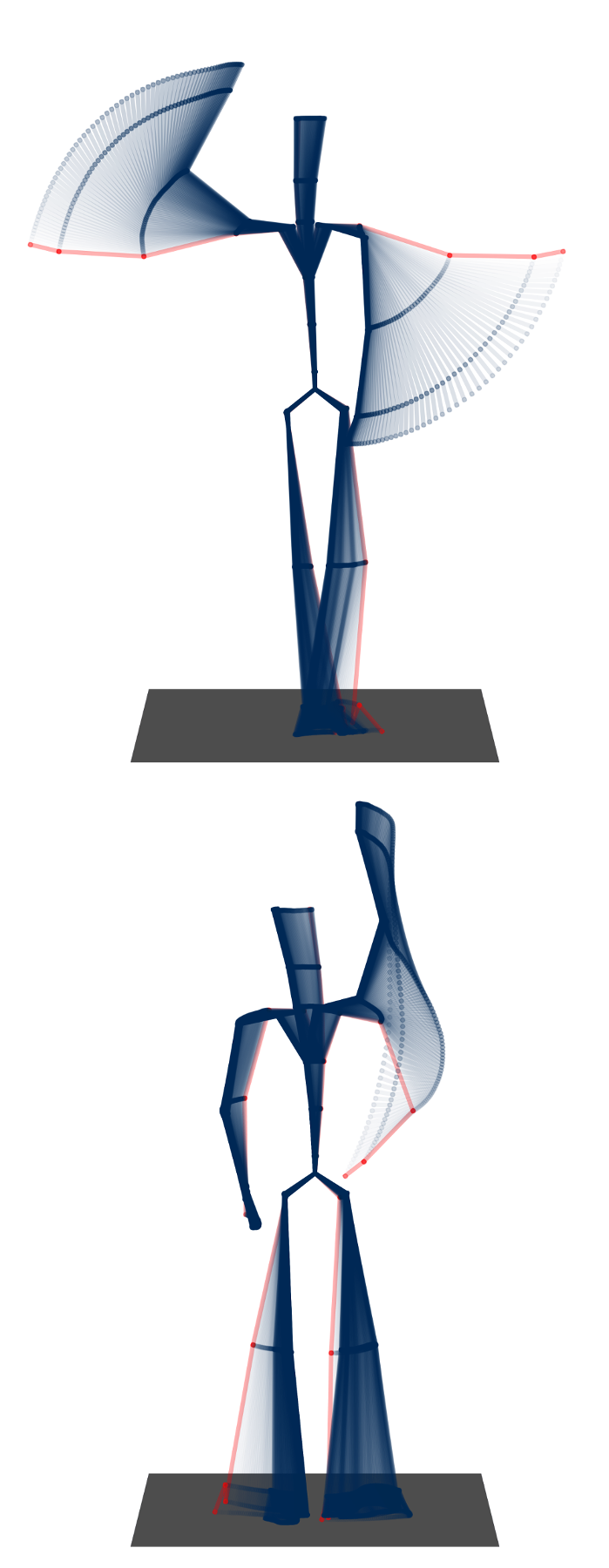}}
    \subfloat[Go. \label{fig:samples_go}]
    {\includegraphics[width=0.19\columnwidth]{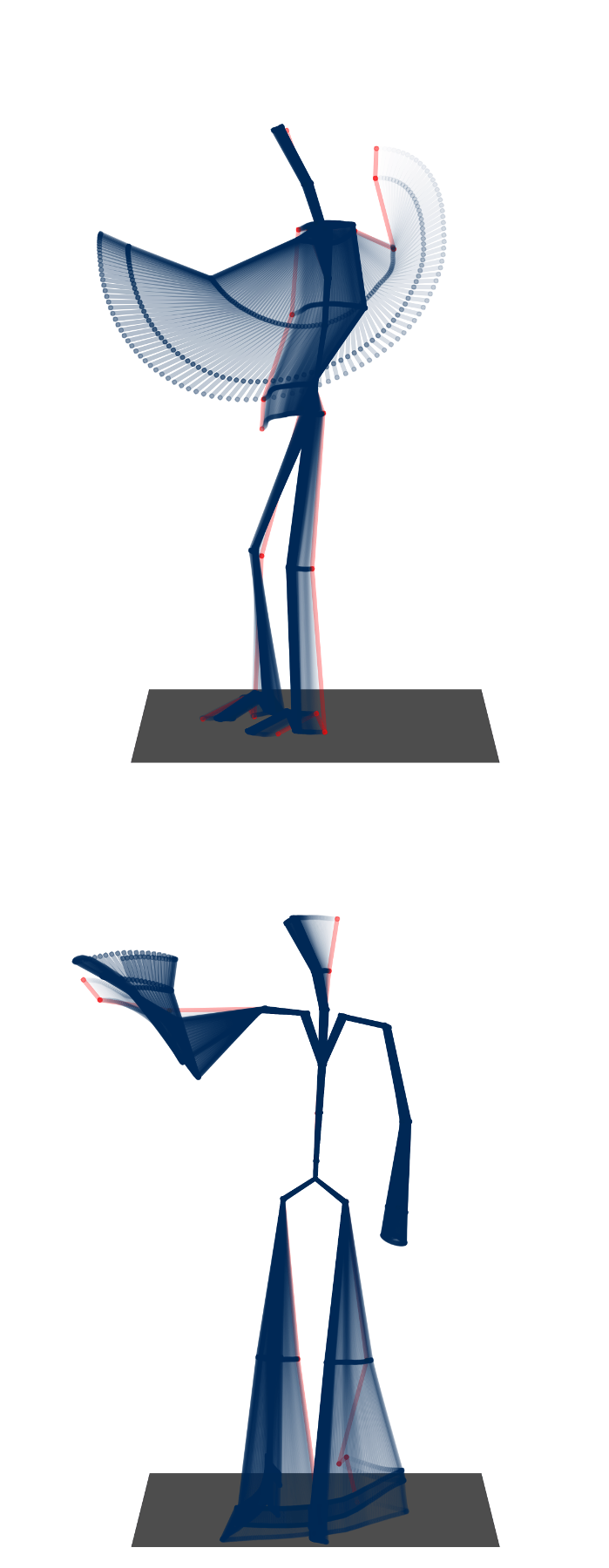}}
    \subfloat[Inactive. \label{fig:samples_inactive}]
    {\includegraphics[width=0.19\columnwidth]{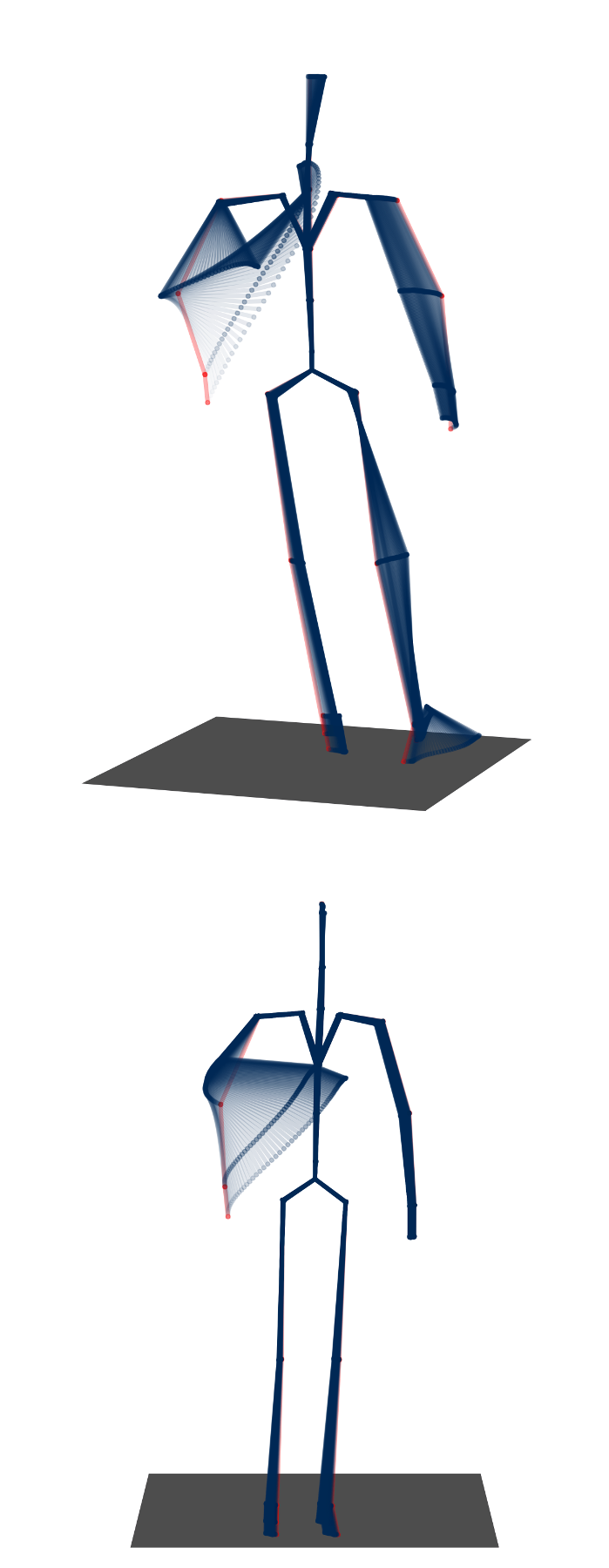}}
    \subfloat[][Skeleton\\Representation. \label{fig:skeleton_model}]
    {\includegraphics[width=0.24\columnwidth, height=0.49\columnwidth]{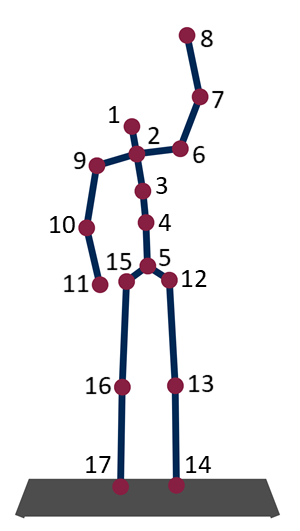}}
    \caption{Our dataset is characterized by high intra-class variance. Fig. \ref{fig:samples_stop} to \ref{fig:samples_inactive} show exceperts of motion sequences, in red the starting pose. In the upper image of Fig. \ref{fig:samples_inactive} the actor is scratching her face and in the lower image someone is looking at his watch. In particular dynamic \emph{go} gestures are challenging \ref{fig:samples_go} due to their similarity to motions from the inactive class.  Fig. (e) describes our 17 joint skeleton model: (1) head, (2) neck, (3) chest, (4) spine, (5) hip, (6) - (11) left shoulder, elbow, hand, thigh, knee, foot and the same for the right-hand side (12) - (17).}
    \label{fig:dataset_samples}
\end{figure}

\subsection{Experimental Settings and Data Collection}

\begin{figure}[t]
    \centering
    \subfloat[Five T-junction scenarios \label{fig:tjunction}]{\includegraphics[width=\columnwidth]{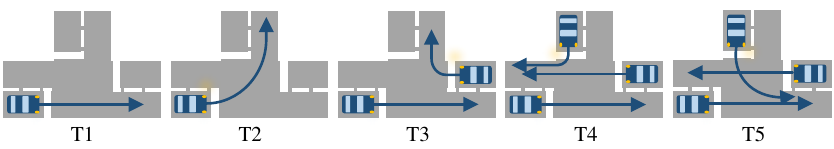}}\\
    \subfloat[Five X-junction scenarios. \label{fig:xjunction}]{\includegraphics[width=\columnwidth]{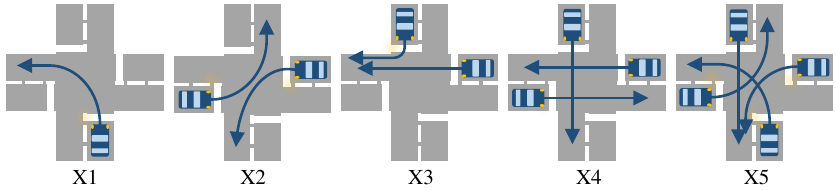}}
    \caption{Birds-eye view on the 10 scenarios of the T- and X-junction. From left to right, we observe the increasing complexity in terms of number of cars, as well as their driving intentions.}
    \label{fig:scenarios}
\end{figure}

\begin{figure}[t]
    \centering
    \subfloat[Subject. \label{fig:pie_subject}]{\includegraphics[width=0.33\columnwidth]{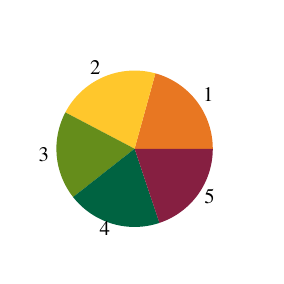}}
    \hfill
    \subfloat[Scenario. \label{fig:pie_scenario}]{\includegraphics[width=0.33\columnwidth]{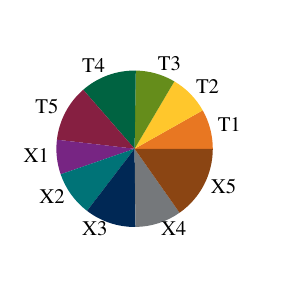}}
    \hfill
    \subfloat[Viewpoint.
    \label{fig:pie_viewpoint}]{\includegraphics[width=0.33\columnwidth]{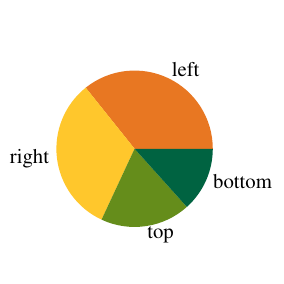}}
    \caption{Frame distribution over subjects (a), scenarios (b) and car viewpoints (c). With the scenario complexity, the sequence length is increasing.}
    \label{fig:frame_distribution}
\end{figure}

We asked from 5 individuals of different body types to regulate the traffic on road intersections. We chose a T- and X-junction where the individual makes uses of the hands for regulation, without additional control devices like whistle or traffic paddle. We also defined 5 different scenarios for each junction, with variable number of involved vehicles. Fig.~\ref{fig:scenarios} shows all scenarios in bird's-eye view, while Fig.~\ref{fig:frame_distribution} presents the data distribution for all scenarios and individuals. The vehicles are specified based on their driving intention, i.e. straight, left turn or right turn, and driving order.

Since staging in real traffic situations is not permitted, we simulate the above scenarios in a closed environment, including intersection layouts, vehicles and the traffic controller. For that reason, we used colored discs to mark the streets and stopping lines. Additional colored markers were placed at the positions of the waiting vehicles to simulate the interaction partners. This helps the actors to adapt their sight according to the marker they are interacting with. In this way the setting facilitates realistic head and body orientations.

To capture the body motion of the traffic controller, each actor has been centered in the road intersection and wore an IMU\footnote{Inertial Measurement Unit (IMU).}-based motion capture suit above the clothes. In total the suit is composed of 17 high-quality MEMS\footnote{Micro Electro Mechanical Systems (MEMS).} inertial sensors (accelerometer, magnetometer and gyroscope) and two pressure insoles to record smooth orientation measurements in high resolution. All sensors were synchronously sampled on 100 Hz and streamed to a computer via integrated Wi-Fi transmitter. Since the computing resources are valuable and limited in an autonomous vehicle, the sampling frequencies can not be very high. For this reason, we sub-sample to 20 Hz as a reasonable frequency for autonomous vehicles. An implemented kinematic body model computes exact 3D locations and orientations of the body joints. In total, we have a skeleton model with 17 3D body joints as it is depicted in Fig.~\ref{fig:dataset_samples}. Of course, the recordings would not be easily feasible outdoors. This is the advantage of the closed environment data collection.

During recordings, a lightweight script helped the actors to keep the correct order of commands, i.e. which car needs to be stopped next and which one should proceed, while they are completely free in the duration of the commands. The script is intended as a high-level guidance rather than a detailed story line, since strong restrictions could lead to insecure and unrealistic behavior. Each scenario is repeated 5 times. In early repetitions we request the actors to perform road traffic control gestures, but after increasing the repetitions, the actors are allowed to use their own, spontaneous gestures in order to control the situation. As a starting point, all actors learn the standard European traffic control gestures, i.e.~stop, go and clear. With this loose recording procedure, we achieve granularity in motion complexity, while the different actors contributed to high motion diversity.

\subsection{Label Definition}

In autonomous driving, the perception provides the environmental state, e.g. object locations or lane markings in each time step. The next action is then planned based on the history and the current state. As a result, traffic control gesture recognition should also happen continuously. To follow this principle, we build our dataset with gesture labels per time step. We reach high annotation quality with trained annotators and consequent quality-checks.

\begin{figure}[t]
    \includegraphics[width=\columnwidth]{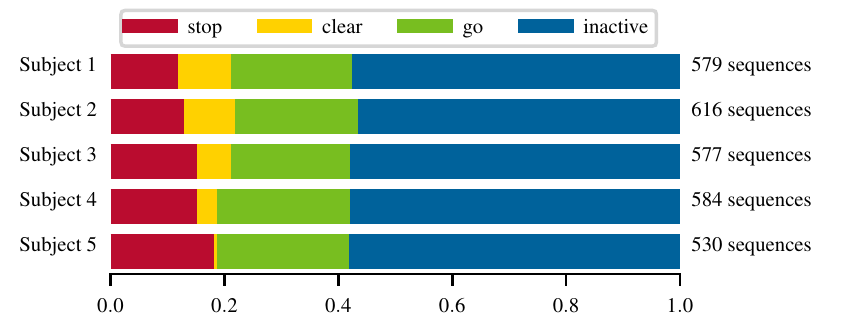}
    \caption{The active classes constitute over 40 \% of the dataset. All classes are well distributed on the 5 actors except for \emph{clear}, which is rarely present for subject 5.}
    \label{fig:class_distribution}
\end{figure}

According to \cite{Gupta2016ConventionalizedVehicles} and the German regulations, we differentiate three active gesture classes, \emph{go}, \emph{clear} and \emph{stop}, as well as an \emph{inactive} class. To increase the diversity of the \emph{inactive} class, we actively enrich motions with daily activities, like rubbing hands, taking sunglasses on or looking at watch. Fig.~\ref{fig:samples_stop} to \ref{fig:samples_inactive} show examples for each class of our dataset. Additionally, the dataset provides annotation for the evaluation of transition phases, e.g. from \emph{Stop} to \emph{Go}. This can give insights for the decision boundaries of the gesture classifier, e.g. a gesture classifier that detects a stop gesture early in the transition phase might be a solution for autonomous driving compared to another one with larger detection latency. For the main classes, \emph{go}, \emph{clear} and \emph{stop}, we sub-categorize the motion of the active hand, e.g. \emph{left}, \emph{right} or \emph{both}, into \emph{static} and \emph{dynamic}. Fig. \ref{fig:class_distribution} compares the class distribution for the 5 subjects with overall 2,886 unique time intervals annotated with a major class label. The \emph{inactive} class dominates over the classes as expected for real traffic situations. Table~\ref{tab:sub_classes} provides quantitative insights of the label distribution. For most of the time, the \emph{go} commands are indicated in a dynamic way, while road traffic controllers signal stop and clear in a more static way. Dynamic stop gestures with both hands are very rare, while dynamic go with a right waving or pointing is highly present. 

\subsection{Dataset Properties}

The dataset includes 250 unique 3D human body pose sequences, ranging from 16 to 90 seconds per sequence. We consider the directional property of gestures. This means that the gesture interpretation strongly depends on the viewpoint. For instance, a static stop gesture from one viewpoint will be a go gesture from another orthogonal viewpoint; or a dynamic go to the right does not mean any signal to the other participants. Therefore, the 3D body poses are transformed in the corresponding coordinate systems of the involved vehicles, i.e. the autonomous vehicle. On average, every sequence is transformed in 2.2 viewpoints, which results in 550 perspectives. 

All sequences are recorded in high temporal resolution of 100 Hz and comprise 140 minutes of realistic human body motion, in total 839,350 frames. As shown in Fig.~\ref{fig:pie_subject}, the amount of frames are evenly distributed on the 5 subjects. Apparently, with the complexity of the scenes, i.e. from \emph{T1} to \emph{T5} and \emph{X1} to \emph{X5}, sequences become longer (Fig.~\ref{fig:pie_scenario}). The pie chart over viewpoints, Fig.~\ref{fig:pie_viewpoint}, shows an under-representation of vehicles coming from the lower street, since it does not appear in the T-junction layout. Based on the design of the scenarios, most of the vehicles approach from the left and right. \\
The proposed TCG dataset can serve the community as a considerable learning base for continuous gesture recognition in the context of self-driving cars.

\begingroup
\setlength{\tabcolsep}{7pt}

\begin{table}[t]
\centering
\begin{tabular}{l|lll}
\textbf{Active Classes} & \textbf{Stop} & \textbf{Clear} & \textbf{Go} \\ \hline
both-hand-static          & 219           & -              & 188                             \\
both-hand-dynamic         & 5             & -              & 8                                \\
left-hand-static          & 157           & 32             & 43                               \\
left-hand-dynamic         & 23            & -              & 179                             \\
right-hand-static         & 133           & 134            & 39                               \\
right-hand-dynamic        & 14            & -              & 339                            \\ \hline
In total             & 551           & 166            & 796                
\end{tabular}
\caption{In total, the TCG dataset contains 1,513 active class annotations with corresponding sub-class labels. Apparently, most of the actors where right-handed, since in more than 60 \% of the one-hand gestures the right hand is used.}
\label{tab:sub_classes}
\end{table}
\endgroup

\section{GESTURE RECOGNITION MODELS}\label{GEModels}

We define hand gesture recognition as sequence modeling, where the input sequence is the track of 3D body pose skeletons $\mathbf{x}_{0}, \dots, \mathbf{x}_{T}$ and the output sequence is the gesture category $\mathbf{y}_{0}, \dots, \mathbf{y}_{T}$. At each time step $t \in T$, the body skeleton $\mathbf{x}_{t} \in \mathcal{R}^{3 \times N}$ is composed of $N$ body joints, represented as a vector. The ground-truth gesture category $\mathbf{y}_{t} \in \mathcal{N}^{K}$ is an one-hot vector of $K$ classes. Our goal is to learn the mapping from the input skeleton to the class category from a set of training data. Without loss of generality, we represent that mapping as:
\begin{equation}
    \mathbf{y}_{0}, \dots, \mathbf{y}_{T} = f(\mathbf{x}_{0}, \dots, \mathbf{x}_{T};\theta)
\end{equation}
where $f:\mathcal{R}^{3 \times N \times T} \rightarrow \mathcal{N}^{K  \times T}$ is the mapping function. We propose to approximate the mapping function based on deep neural networks. We consider recurrent, temporal convolutional and graph convolution neural networks as three different ways to approach the problem. For all network architectures, the learning goal is to minimize the difference between the predictions and ground-truth. This can be formalized by the loss function that is given by:
\begin{equation}
\operatorname*{argmin}_\theta  L(\mathbf{y}_{0}, \dots, \mathbf{y}_{T}, f(\mathbf{x}_{0}, \dots, \mathbf{x}_{T};\theta)),
\end{equation}
that is cross-entropy for problem. Finally, the training is accomplished with back-propagation and stochastic gradient descent. Note, that we do not assume access to future time steps, i.e.~$T+1$. Next, we comment on the neural network models for each architecture type.

\begin{figure*}[t]
    \centering
    \includegraphics[width=0.9\textwidth]{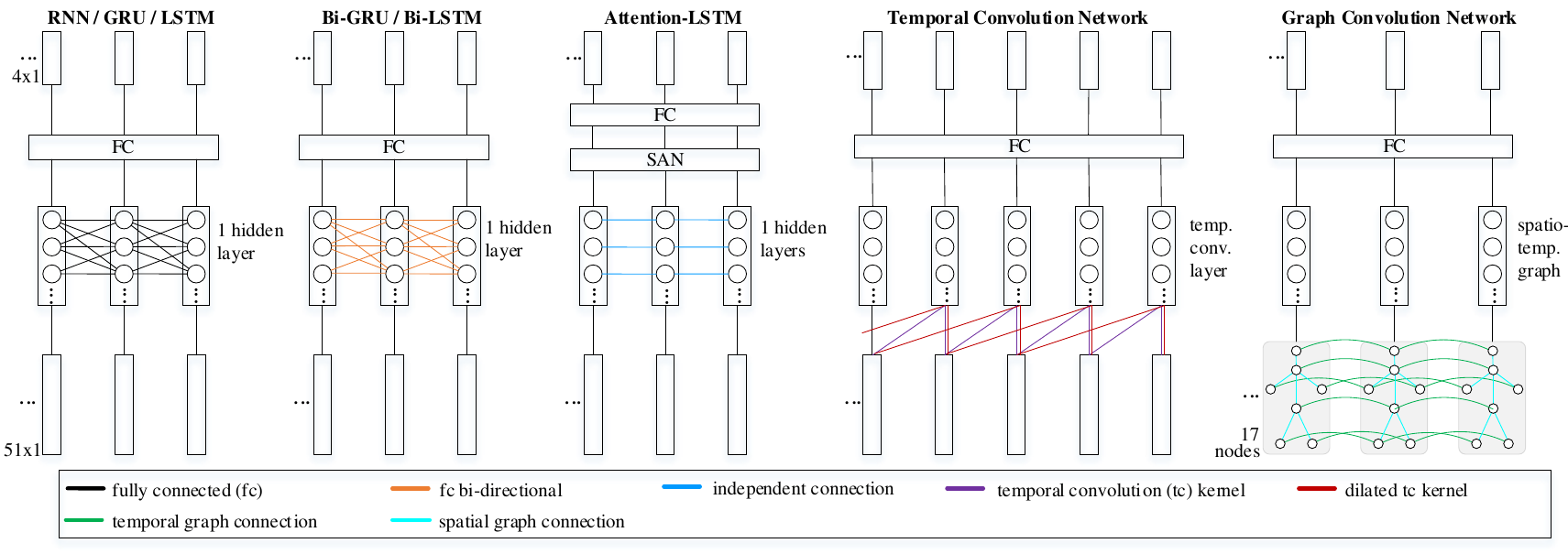}
    \caption{ Network Architectures Illustration. We show the structure of the RNN, GRU and LSTM as well as temporal and graph convolutional networks and their connectivity. The input vector for each time step is the 3D skeleton represented by 17 body joints. We refer to fully connected layer and self-attention networks as FC and SAN, respectively.}
    \label{ModelsArch}
\end{figure*}

\subsection{Recurrent Network Architectures}

Skeleton-based action recognition approaches traditionally make use of RNNs to model the temporal dynamics. GRU\nobreakseq{-,} LSTM-cells or more complex structures, such as bidirectional networks~\cite{Zhu2016Co-OccurrenceNetworks} are the common network architectures since vanilla RNNs do not capture long dependencies. In our evaluation, we consider all these types of RNNs for gesture recognition. 

\subsection{Attention Mechanism}

Modeling long sequences can be accomplished with an attention mechanism as well. Song \textit{et al.}~\cite{Song2017AnData} have shown an end-to-end spatial and temporal attention model for human action recognition. The model is trained to pay more attention on discriminative joints of the skeleton within each frame and to estimate the importance of frames in the sequence. Attention has also been used for spatiotemporal attention networks to model the evolution of dynamic hand gestures~\cite{Hou2019Spatial-temporalRecognition}. To retrieve a better semantic information, a novel model with self-attention network (SAN) was proposed by \cite{Cho2019Self-AttentionRecognition}. We also examine the potential of self-attention in combination with the LSTM cells.

\subsection{Temporal Convolutional Networks}

Recently, it has been shown that convolutional network architectures are on par with recurrent networks on sequence modeling~\cite{bai2018empirical}. At the same time, the idea of temporal convolutions has been established for visual tasks~\cite{Lea2017TemporalDetection}, audio generation~\cite{oord2016wavenet} and signal processing~\cite{casas2018adversarial}. We study the effect of temporal convolutions in our problems as well. The temporal convolutions process the 3D body joints, independently, over-time.

\subsection{Graph Convolutional Architectures}

Graph neural networks are well-suited to non-structured data such as the human body, represented by a skeleton model~\cite{Si2019AnRecognition}. Yan \textit{et al.}~\cite{Li2019SpatialRecognition} proposed a spatiotemporal graph convolutional network to perform activity recognition from skeletal data. The skeletons are composed of 2D or 3D joint positions. We rely on the same idea to perform gesture recognition. We present a graph convolutional network that processes 3D body joints to classify traffic control activities.

\begin{table*}[t]
\centering
\caption{Results on the 4-Class Evaluation. We perform cross-subject, cross-view and real-world evaluations for all models and provide the mean and standard deviation of three runs. For all metrics, the higher score the better the result.}
\resizebox{\textwidth}{!}{\begin{tabular}{|l|ccc|ccc|ccc|}
\multicolumn{1}{|c|}{\multirow{2}{*}{\textbf{Methods}}} & \multicolumn{3}{c|}{\textbf{Cross-subject}} & \multicolumn{3}{c|}{\textbf{Cross-view}} & \multicolumn{3}{c|}{\textbf{Real-World}} \\
\multicolumn{1}{|c|}{}                                  & Accuracy      & Jaccard      & F1-score     & Accuracy     & Jaccard     & F1-score    & Accuracy     & Jaccard     & F1-score    \\ \hline
RNN \cite{Shahroudy2016NTUAnalysis}
& 82.81 (\textit{\scriptsize$\pm$\scriptsize 2.7}) 
& 57.40 (\textit{\scriptsize$\pm$\scriptsize 2.3}) 
& 69.45 (\textit{\scriptsize$\pm$\scriptsize 1.4}) 
& 80.94 (\textit{\scriptsize$\pm$\scriptsize 1.9}) 
& 57.21 (\textit{\scriptsize$\pm$\scriptsize 2.5}) 
& 69.98 (\textit{\scriptsize$\pm$\scriptsize 2.3})
& 69.39 (\textit{\scriptsize$\pm$\scriptsize 7.2}) 
& 39.70 (\textit{\scriptsize$\pm$\scriptsize 8.6})
& 50.26 (\textit{\scriptsize$\pm$\scriptsize 10.2})\\     

GRU      
& 84.44 (\textit{\scriptsize$\pm$\scriptsize 2.0}) 
& 58.16 (\textit{\scriptsize$\pm$\scriptsize 4.2}) 
& 70.45 (\textit{\scriptsize$\pm$\scriptsize 3.1})
& 83.47 (\textit{\scriptsize$\pm$\scriptsize 1.4}) 
& 56.25 (\textit{\scriptsize$\pm$\scriptsize 7.6}) 
& 68.59 (\textit{\scriptsize$\pm$\scriptsize 7.4})
& 71.8 (\textit{\scriptsize$\pm$\scriptsize 8.6}) 
& 40.4 (\textit{\scriptsize$\pm$\scriptsize 10.2}) 
& 50.67 (\textit{\scriptsize$\pm$\scriptsize 11.4})    \\

LSTM  \cite{Shahroudy2016NTUAnalysis}     
& 83.23 (\textit{\scriptsize$\pm$\scriptsize3.6}) 
& 56.32 (\textit{\scriptsize$\pm$\scriptsize 7.0})  
& 68.59 (\textit{\scriptsize$\pm$\scriptsize 6.9})
& 79.58 (\textit{\scriptsize$\pm$\scriptsize 1.6}) 
& 52.02 (\textit{\scriptsize$\pm$\scriptsize 3.2}) 
& 64.62 (\textit{\scriptsize$\pm$\scriptsize 3.8})
& \textbf{77.88 (\textit{\scriptsize$\pm$\scriptsize 9.6})} 
& \textbf{52.90 (\textit{\scriptsize$\pm$\scriptsize 15.0})} 
& \textbf{62.21 (\textit{\scriptsize$\pm$\scriptsize 15.2})} \\

Att-LSTM  
& 85.67 (\textit{\scriptsize$\pm$\scriptsize2.1}) 
& 50.70 (\textit{\scriptsize$\pm$\scriptsize 9.9})
& 61.87 (\textit{\scriptsize$\pm$\scriptsize 10.6})
& 85.30 (\textit{\scriptsize$\pm$\scriptsize 1.1})
& 59.87 (\textit{\scriptsize$\pm$\scriptsize 12.7})
& 71.20 (\textit{\scriptsize$\pm$\scriptsize 12.3})
& 72.76 (\textit{\scriptsize$\pm$\scriptsize 10.2})
& 44.61 (\textit{\scriptsize$\pm$\scriptsize 15.4})
& 52.50 (\textit{\scriptsize$\pm$\scriptsize 16.0})        \\

Bi-GRU   
& 86.80 (\textit{\scriptsize$\pm$\scriptsize1.6}) 
& 57.25 (\textit{\scriptsize$\pm$\scriptsize 7.4}) 
& 68.95 (\textit{\scriptsize$\pm$\scriptsize 6.4})
& \textbf{87.37 (\textit{\scriptsize$\pm$\scriptsize 0.3})} 
& 55.55 (\textit{\scriptsize$\pm$\scriptsize 2.8}) 
& 67.68 (\textit{\scriptsize$\pm$\scriptsize 2.2})
& 73.58 (\textit{\scriptsize$\pm$\scriptsize 8.1}) 
& 43.09 (\textit{\scriptsize$\pm$\scriptsize 10.8}) 
& 52.26 (\textit{\scriptsize$\pm$\scriptsize 12.8}) \\

Bi-LSTM \cite{Zou2019DeepBiLSTMRecognition}     
& \textbf{87.24 (\textit{\scriptsize$\pm$\scriptsize1.8})} 
& \textbf{67.00 (\textit{\scriptsize$\pm$\scriptsize 2.1})} 
& \textbf{78.48 (\textit{\scriptsize$\pm$\scriptsize 1.8})}
& 86.66 (\textit{\scriptsize$\pm$\scriptsize 1.2)}
& \textbf{65.95 (\textit{\scriptsize$\pm$\scriptsize 4.7})} 
& \textbf{77.14 (\textit{\scriptsize$\pm$\scriptsize 4.3})}
& 72.28 (\textit{\scriptsize$\pm$\scriptsize 8.7)}
& 48.81 (\textit{\scriptsize$\pm$\scriptsize 12.5)}
& 61.23 (\textit{\scriptsize$\pm$\scriptsize 14.4)} \\

TCN \cite {Lea2017TemporalDetection}
& 83.44 (\textit{\scriptsize$\pm$\scriptsize2.5}) 
& 62.06 (\textit{\scriptsize$\pm$\scriptsize 2.8})
& 74.23 (\textit{\scriptsize$\pm$\scriptsize 3.0})
& 82.66 (\textit{\scriptsize$\pm$\scriptsize 0.7})
& 63.97 (\textit{\scriptsize$\pm$\scriptsize 1.3}) 
&  75.95 (\textit{\scriptsize$\pm$\scriptsize 0.9})
& 48.70 (\textit{\scriptsize$\pm$\scriptsize 6.5}) 
& 25.73 (\textit{\scriptsize$\pm$\scriptsize 6.4}) 
& 35.23 (\textit{\scriptsize$\pm$\scriptsize 8.2}) \\

GCN \cite{Li2019SpatialRecognition}
& 65.42 (\textit{\scriptsize$\pm$\scriptsize9.6}) 
& 38.55 (\textit{\scriptsize$\pm$\scriptsize 13.6}) 
& 50.73 (\textit{\scriptsize$\pm$\scriptsize 14.5})
& 62.40 (\textit{\scriptsize$\pm$\scriptsize 14.2}) 
& 35.05 (\textit{\scriptsize$\pm$\scriptsize 9.8})  
& 48.51 (\textit{\scriptsize$\pm$\scriptsize 11.3})
& 60.64 (\textit{\scriptsize$\pm$\scriptsize 3.7})
& 34.34 (\textit{\scriptsize$\pm$\scriptsize 2.9})
& 48.72 (\textit{\scriptsize$\pm$\scriptsize 3.4})
\end{tabular}}
\label{EvalDataset4Class}
\end{table*}

\begin{figure*}[t]
\centering
\subfloat[Vanilla-RNN \label{fig:rnn_xs}]{\includegraphics[width=0.125\textwidth]{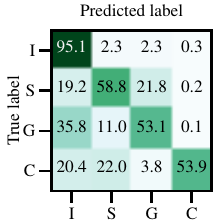}}
\hfill
\subfloat[GRU \label{fig:gru_xs}]{\includegraphics[width=0.125\textwidth]{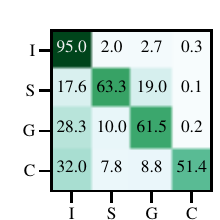}}
\hfill
\subfloat[LSTM \label{fig:lstm_xs}]{\includegraphics[width=0.125\textwidth]{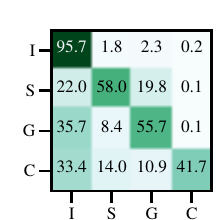}}
\hfill
\subfloat[Att-LSTM \label{fig:attlstm_xs}]{\includegraphics[width=0.125\textwidth]{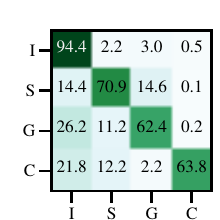}}
\hfill
\subfloat[Bi-GRU \label{fig:bigru_xs}]{\includegraphics[width=0.125\textwidth]{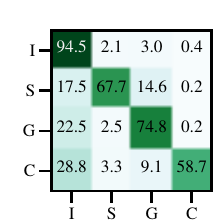}}
\hfill
\subfloat[Bi-LSTM \label{fig:bilstm_xs}]{\includegraphics[width=0.125\textwidth]{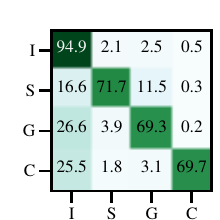}}
\hfill
\subfloat[TCN \label{fig:tcn_xs}]{\includegraphics[width=0.125\textwidth]{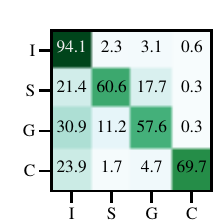}}
\hfill
\subfloat[GCN \label{fig:gcn_xs}]{\includegraphics[width=0.125\textwidth]{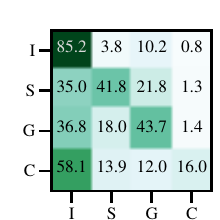}}
\caption{Cross-Subject Confusion Matrices on 4-Class. We abbreviate the gestures \emph{inactive}, \emph{stop}, \emph{go} \& \emph{clear} as \emph{I}, \emph{S}, \emph{G} \& \emph{C}.}
\label{CrossSubConf}
\end{figure*}

\begin{figure*}[t]
\centering
\subfloat[Vanilla-RNN \label{fig:rnn_xv}]{\includegraphics[width=0.125\textwidth]{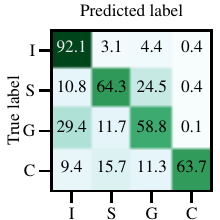}}
\hfill
\subfloat[GRU \label{fig:gru_xv}]{\includegraphics[width=0.125\textwidth]{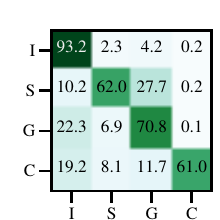}}
\hfill
\subfloat[LSTM \label{fig:lstm_xv}]{\includegraphics[width=0.125\textwidth]{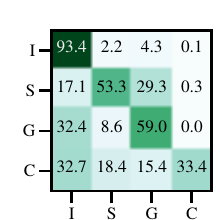}}
\hfill
\subfloat[Att-LSTM \label{fig:attlstm_xv}]{\includegraphics[width=0.125\textwidth]{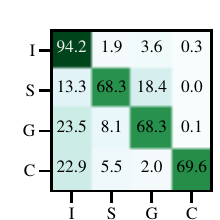}}
\hfill
\subfloat[Bi-GRU \label{fig:bigru_xv}]{\includegraphics[width=0.125\textwidth]{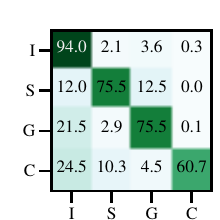}}
\hfill
\subfloat[Bi-LSTM \label{fig:bilstm_xv}]{\includegraphics[width=0.125\textwidth]{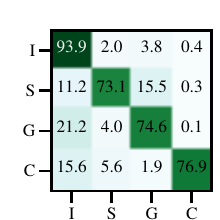}}
\hfill
\subfloat[TCN \label{fig:tcn_xv}]{\includegraphics[width=0.125\textwidth]{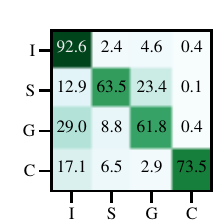}}
\hfill
\subfloat[GCN \label{fig:gcn_xv}]{\includegraphics[width=0.125\textwidth]{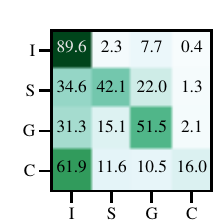}}
\caption{Cross-View Confusion Matrices on 4-Class. We abbreviate the gestures \emph{inactive}, \emph{stop}, \emph{go} \& \emph{clear} as \emph{I}, \emph{S}, \emph{G} \& \emph{C}.}
\label{CrossViewConf}
\end{figure*}

\section{EXPERIMENTS}

We evaluate the presented dataset for different sequence modelling strategies, as they have been presented in Sec.~\ref{GEModels}. The experiments include six recurrent network models, one temporal convolution network (TCN) and a spatio-temporal graph convolution network (GCN). Similar to gesture recognition approaches~\cite{Shahroudy2016NTUAnalysis}, the evaluation metrics are accuracy, as well as Jaccard index~\cite{Escalera2015ChalearnResults}, F1-score and the confusion matrix. At last, we present an image-based evaluation on real-world sequences with the traffic officer and the autonomous vehicle.

\subsection{Network Architecture Implementation}

We provide the implementation details for each neural network model individually. In general, all models have been trained from scratch with grid hyper-parameter search. Moreover, the activation function is non-linear, dropout is applied everywhere with rate 0.5 and the training takes place until convergence. Class confidences are computed with a dense layer and the softmax function on top of the high-level feature representations provided by the temporal models. The optimizer is the adaptive learning rate optimization algorithm (Adam)~\cite{kingma2014method}, with initial learning rate 0.001, unless it is differently reported. Below, the specific configuration for each temporal model is reported.

\paragraph{Recurrent Neural Networks}

We consider six types of recurrent neural networks, combined with a fully connected layer and the softmax activation function to perform gesture classification. In detail, the encoder is modeled as \emph{vanilla-RNN}, \emph{GRU}, \emph{LSTM}, \emph{bidirectional-GRU} or \emph{bidirectional-LSTM}. Since the sequence length varies, we adopt a masking mechanism for the input 3D body skeletons as in~\cite{Zou2019DeepBiLSTMRecognition} to overcome the zero-padding problem. For the bidirectional-LSTM, we adopt the architecture of~\cite{Zou2019DeepBiLSTMRecognition}. For the other models, our architecture is presented in Fig.~\ref{ModelsArch}. In all cases, we rely on 100 cells and a single hidden layer.

\paragraph{Attention Model}

We add an attention layer on top of the \emph{LSTM} encoder. In particular, we transform the \emph{LSTM} to \emph{Attention-LSTM} and make use of same architecture as before, however, empirically select 50 cells for the hidden layer and 50 attention units.

\paragraph{Temporal Convolutional Networks}

We adopt~\cite{Lea2017TemporalDetection} to implement our TCN. We build it though simpler, because it does not deal with image data. It consists of 1D convolution kernels of size 2 and 64 filters where the dilation rate goes from 2 and reaches 32, by doubling it in each layer. The parameter optimization is accomplished with Adam, with learning rate 0.001 and back-propagation. An illustration of the model is shown in Fig.~\ref{ModelsArch}.

\paragraph{Graph Convolutional Networks}

We rely on the GCN of~\cite{Li2019SpatialRecognition} for our problem. The body is represented as an undirected spatiotemporal graph with 17 joints and T time steps, where T=20 for making a single prediction. In the training phase we randomly sample sequences of 20 3D body skeletons of each class. During testing continuous predictions are required. Therefore we predict with a sliding window of stride 1 to obtain continuous predictions that equally compare with the other sequence modelling approaches. The initial learning rate for this model is 0.1.

\subsection{Cross-Subject \& Cross-View Protocol}

We define the cross-subject and cross-view evaluation protocol, similar to the gesture recognition approaches~\cite{Shahroudy2016NTUAnalysis}. Note we explicitly aim to distinguish gestures dependent on a specific viewpoint, i.e. view variant recognition. In the cross-view evaluation, this means that the labels differ depending on the vehicle's viewpoint. As a result, the model is trained on all sequences of 3 viewpoints, e.g left, top and right, and evaluated on the omitted set of sequences, e.g. bottom. In the cross-subject evaluation, which is considered to be more challenging, the model is trained on 4 actors and tested on the remaining actor. The process is repeated for all combinations.

\subsection{Real-World Experiment Description}\label{realwroldDesc}

Our ultimate goal is to make use of our dataset for real-world traffic control gesture recognition. For that reason, we captured 5 image-based sequences of real traffic control scenarios. They consist of the traffic regulator on a T-junction road intersection and the autonomous vehicle. After labeling the sequences, we perform an evaluation of the presented gesture recognition approaches.

To obtain the 3D body skeleton from the image input, first, the traffic regulator is detected and the body 2D pose is extracted with a pre-trained Mask-RNN~\cite{He2017} model. Since the 3D pose is necessary, we rely on the approach of Pavllo \textit{et al.}~\cite{Pavllo20183DTraining} to lift the 2D body poses to 3D body pose skeletons based on a sequence of 2D poses. Second, the estimated 3D body pose skeletons are provided to the gesture recognition approach for classification. We have performed this experiment off-line for being able to follow our evaluation protocol. Our approach is outlined in Fig.~\ref{fig:real_world_demo}.

\begin{table*}[]
\centering
\caption{Results on the 15-Class Evaluation. We perform cross-subject, cross-view and real-world evaluations for all models and provide the mean and standard deviation of three runs. For all metrics, the higher score the better the result. We skip the results of the GCN because of poor performance.}
\resizebox{\textwidth}{!}{\begin{tabular}{|l|ccc|ccc|ccc|}
\multicolumn{1}{|c|}{\multirow{2}{*}{\textbf{Methods}}} & \multicolumn{3}{c|}{\textbf{Cross-subject}} & \multicolumn{3}{c|}{\textbf{Cross-view}} & \multicolumn{3}{c|}{\textbf{Real-World}} \\
\multicolumn{1}{|c|}{}                                  & Accuracy      & Jaccard      & F1-score     & Accuracy     & Jaccard     & F1-score    & Accuracy     & Jaccard     & F1-score    \\ \hline

RNN \cite{Shahroudy2016NTUAnalysis}
& 78.44 (\textit{\scriptsize$\pm$\scriptsize 1.2})
& 19.19 (\textit{\scriptsize$\pm$\scriptsize 3.1}) 
& 25.33 (\textit{\scriptsize$\pm$\scriptsize 3.8})
& 80.84 (\textit{\scriptsize$\pm$\scriptsize 1.0})
& 24.39 (\textit{\scriptsize$\pm$\scriptsize 2.6}) 
& 31.00 (\textit{\scriptsize$\pm$\scriptsize 2.8})
& 71.28 (\textit{\scriptsize$\pm$\scriptsize 8.4}) 
& 30.29 (\textit{\scriptsize$\pm$\scriptsize 19.8}) 
& 33.12 (\textit{\scriptsize$\pm$\scriptsize 19.0})\\

GRU   
& 79.27 (\textit{\scriptsize$\pm$\scriptsize 1.0}) 
& 28.59 (\textit{\scriptsize$\pm$\scriptsize 4.1}) 
& 36.09 (\textit{\scriptsize$\pm$\scriptsize 4.7})
& 81.58 (\textit{\scriptsize$\pm$\scriptsize 0.7}) 
& 26.30 (\textit{\scriptsize$\pm$\scriptsize 0.9}) 
& 33.74 (\textit{\scriptsize$\pm$\scriptsize 1.3})
& 73.08 (\textit{\scriptsize$\pm$\scriptsize 13.2}) 
& 30.11 (\textit{\scriptsize$\pm$\scriptsize 12.0}) 
& 31.55 (\textit{\scriptsize$\pm$\scriptsize 12.1}) \\

LSTM  \cite{Shahroudy2016NTUAnalysis}
& 73.26 (\textit{\scriptsize$\pm$\scriptsize 1.7}) 
& 17.88 (\textit{\scriptsize$\pm$\scriptsize 2.9}) 
& 22.26 (\textit{\scriptsize$\pm$\scriptsize 3.2})
& 73.31 (\textit{\scriptsize$\pm$\scriptsize 0.6}) 
& 12.98 (\textit{\scriptsize$\pm$\scriptsize 1.3}) 
& 16.71 (\textit{\scriptsize$\pm$\scriptsize 1.7})
& \textbf{75.62 (\textit{\scriptsize$\pm$\scriptsize 8.7})}
& \textbf{40.42 (\textit{\scriptsize$\pm$\scriptsize 14.2})} 
& \textbf{45.14 (\textit{\scriptsize$\pm$\scriptsize 13.1})} \\

Att-LSTM
& 79.90 (\textit{\scriptsize$\pm$\scriptsize 1.3})
& 22.92 (\textit{\scriptsize$\pm$\scriptsize 3.9}) 
& 29.91 (\textit{\scriptsize$\pm$\scriptsize 3.9})
& 83.49 (\textit{\scriptsize$\pm$\scriptsize 0.8}) 
& 23.01 (\textit{\scriptsize$\pm$\scriptsize 3.9}) 
& 30.73 (\textit{\scriptsize$\pm$\scriptsize 4.3})
& 71.96 (\textit{\scriptsize$\pm$\scriptsize 16.2}) 
& 32.06 (\textit{\scriptsize$\pm$\scriptsize 11.7}) 
& 35.04 (\textit{\scriptsize$\pm$\scriptsize 11.2}) \\

Bi-GRU
& \textbf{82.70 (\textit{\scriptsize$\pm$\scriptsize 1.1})}
& 27.8 (\textit{\scriptsize$\pm$\scriptsize 4.8})
& 35.9 (\textit{\scriptsize$\pm$\scriptsize 5.0})
& 83.59 (\textit{\scriptsize$\pm$\scriptsize 0.9}) 
& 25.9 (\textit{\scriptsize$\pm$\scriptsize 3.7}) 
& 33.56 (\textit{\scriptsize$\pm$\scriptsize 4.0})
& 75.17 (\textit{\scriptsize$\pm$\scriptsize 11.7}) 
& 30.6 (\textit{\scriptsize$\pm$\scriptsize 20.5}) 
& 33.14 (\textit{\scriptsize$\pm$\scriptsize 19.8})
 \\

Bi-LSTM \cite{Zou2019DeepBiLSTMRecognition}
& 82.46 (\textit{\scriptsize$\pm$\scriptsize 0.9}) 
& \textbf{29.42 (\textit{\scriptsize$\pm$\scriptsize 4.7})} 
& \textbf{37.77 (\textit{\scriptsize$\pm$\scriptsize 5.3})}
& \textbf{84.27 (\textit{\scriptsize$\pm$\scriptsize 1.0})} 
& \textbf{27.76 (\textit{\scriptsize$\pm$\scriptsize 2.6})} 
& \textbf{35.74 (\textit{\scriptsize$\pm$\scriptsize 2.7})}
& 71.75 (\textit{\scriptsize$\pm$\scriptsize 12.4})
& 30.80 (\textit{\scriptsize$\pm$\scriptsize 20.9}) 
& 33.75 (\textit{\scriptsize$\pm$\scriptsize 20.1}) \\

TCN \cite {Lea2017TemporalDetection}
& 73.17 (\textit{\scriptsize$\pm$\scriptsize 3.8}) 
& 11.55 (\textit{\scriptsize$\pm$\scriptsize 5.9})
& 15.19 (\textit{\scriptsize$\pm$\scriptsize 7.9})
& 74.84 (\textit{\scriptsize$\pm$\scriptsize 2.2}) 
& 15.09 (\textit{\scriptsize$\pm$\scriptsize 4.9}) 
& 19.09 (\textit{\scriptsize$\pm$\scriptsize 6.8})
& 65.80 (\textit{\scriptsize$\pm$\scriptsize 8.8})
& 29.75 (\textit{\scriptsize$\pm$\scriptsize 8.4})
& 32.21 (\textit{\scriptsize$\pm$\scriptsize 12.4})\\
\end{tabular}}
\label{EvalDataset15Class}
\end{table*}

\subsection{Quantitative Evaluation}\label{QauntEval}

The dataset evaluation is  performed for the 4-class problem, i.e.~\emph{go}, \emph{clear}, \emph{stop} and \emph{inactive}. Both training and test sets come from our dataset according to the cross-subject and cross-view protocol. For the real-world evaluation, the test set is the image-based real-world sequences, as described in Sec.~\ref{realwroldDesc}. Since one actor of the dataset also appears in the real-world image sequences, we exclude the actor from the dataset and re-train all models. The dataset results for cross-subject, cross-view, as well as the real-world evaluation are presented in Table~\ref{EvalDataset4Class}. Especially in unbalanced recognition tasks, a fair metric is required to take the distribution of classes into account. For that reason we consider the Jaccard index as the most representative metric~\cite{Escalera2015ChalearnResults}.

\paragraph{Cross-Subject \& Cross-View} The three evaluation metrics have similar behaviour for the cross-subject and cross-view. The best performing approach is the LSTM in the bidirectional formulation for both cases as shown in Table~\ref{EvalDataset4Class}. Only, the accuracy of bidirectional-GRU is slightly higher than bidirectional-LSTM for the cross-view evaluation. The recurrent networks have in overall comparable performance other than the vanilla RNN. The temporal convolutional network has consistent results both for cross-subject and cross-view, but it is behind the recurrent models. At last, the graph convolutional network has much lower performance compared to all other models. In addition, it had difficulties to converge. We additionally provide the confusion matrices for the cross-subject (Fig.~\ref{CrossSubConf}) and cross-view (Fig.~\ref{CrossViewConf}) evaluation. All classifiers are able to distinct the active classes from the inactive class. Notable is the performance on \emph{go} compared to \emph{stop}. In most of the cases, the recognition performance is higher on the latter, which we explain with the larger amount of dynamic gestures in the \emph{go} class.

\paragraph{Real-World} For the real-world evaluation, all metrics agree on the best performing approach as well. The LSTM model delivers the best results on the real-world sequences (see Table~\ref{EvalDataset4Class}), while here the bidirectional formulation does not further improve the final outcome. Next, the behavior of the temporal convolutional network is similar to the cross-subject and cross-view evaluation. In total, the real-world evaluation delivers considerable worse performance than the cross-view and cross-subject evaluation. This is expected given that the 3D body pose skeletons are algorithmically computed and thus include some sort of error.

\subsection{Ablation Study}

We consider another classification scheme of 15-class\footnote{15-Classes: \emph{inactive}; stop: \emph{both-static}, \emph{both-dynamic}, \emph{left-static}, \emph{left-dynamic}, \emph{right-static}, \emph{right-dynamic}; clear: \emph{left-static}, \emph{right-static}; go: \emph{both-static}, \emph{both-dynamic}, \emph{left-static}, \emph{left-dynamic}, \emph{right-static}, \emph{right-dynamic}.} problem. By moving from 4 to 15 gesture categories, our aim is to study how the static and dynamic gestures affect the classification performance. All experimental settings are the same with Sec.~\ref{QauntEval} except the loss function that is optimized for 15 classes. The results are reported in Table~\ref{EvalDataset15Class}. We report the results of all methods except the graph convolutional network because it has shown unstable convergence during training and thus reached poor performance.

\paragraph{Cross-Subject \& Cross-View} The accuracy for cross-subject and cross-view is comparable to the 4-class problem. However, the jaccard index and F1-score show that the 15-class problem results in a descent performance reduction. This observation holds for all models. The best performing model is again the bidirectional-LSTM for both evaluations. The bidirectional-GRU accuracy is the best for the cross-subject evaluation, but the bidirectional-LSTM is in general on par with it. The other model have similar behavior by comparing the 4-class results of Table~\ref{EvalDataset4Class} with Table~\ref{EvalDataset15Class}.

\paragraph{Real-World} Unlike the cross-subject and cross-view results, the real-world performance is similar to the 4-class problem (see Table~\ref{EvalDataset15Class}). Considering the standard deviation, all models show great variation between runs as a result of the domain difference between the motion capture data for training and the estimated 3D poses for testing. The clear observation is that the LSTM model delivers promising performance.

\section{CONCLUSION}

We introduced a road traffic control gesture recognition dataset in the context of autonomous driving. Our dataset consists of 3D body skeleton data and gesture category for every time step. To perform gesture classification, we presented eight sequential processing models based on deep neural networks, such as recurrent networks, temporal convolutional networks and graph convolutional networks. Finally, we demonstrated promising performance on real-world sequences, which indicates the representativity for our dataset.

\section*{ACKNOWLEDGMENT}
Part of the research was conducted within @CITY-AF (Research project No. 19 A 18003 A.), funded by BMWi (Federal Ministry for Economic Affairs and Energy).

\bibliographystyle{IEEEtran}
\bibliography{all_refs.bib}

\end{document}